\documentclass[sigconf,screen]{acmart}
\AtBeginDocument{%
  }

\setcopyright{none}

\usepackage{tabulary}
\usepackage{hyperref}

\usepackage{float}

\acmConference[C+J '24]{Computation+Journalism Symposium}{October 25--27,
  2024}{Boston, MA}
\acmISBN{}
\acmDOI{}

\usepackage[frozencache,cachedir=.]{minted}
\definecolor{bg}{rgb}{0.95,0.95,0.95}
\begin{document}

\title{De-jargonizing Science for Journalists with GPT-4: A Pilot Study}

\author{Sachita Nishal}
\affiliation{%
 \institution{Northwestern University}
 \country{Evanston, USA}
}
\email{nishal@u.northwestern.edu}

\author{Eric Lee}
\affiliation{%
 \institution{Northwestern University}
 \country{Evanston, USA}
}
\email{ericlee2026@u.northwestern.edu}

\author{Nicholas Diakopoulos}
\affiliation{%
 \institution{Northwestern University}
 \country{Evanston, USA}
}
\email{nad@u.northwestern.edu}

\renewcommand{\shortauthors}{Nishal et al. 2024}

\begin{abstract}
This study offers an initial evaluation of a human-in-the-loop system leveraging GPT-4 (a large language model or LLM), and Retrieval-Augmented Generation (RAG) to identify and define jargon terms in scientific abstracts, based on readers' self-reported knowledge. The system achieves fairly high recall in identifying jargon and preserves relative differences in readers' jargon identification, suggesting personalization as a feasible use-case for LLMs to support sense-making of complex information. Surprisingly, using only abstracts for context to generate definitions yields slightly more accurate and higher quality definitions than using RAG-based context from the fulltext of an article. The findings highlight the potential of generative AI for assisting science reporters, and can inform future work on developing tools to simplify dense documents.

\end{abstract}


\maketitle

\section{Introduction}

Science reporters play a crucial role in translating complex scientific information for the public, helping readers make informed decisions, supporting watchdogging activities, and advocating for policy change. However, reporters face significant challenges when sourcing stories from various repositories of information, such as from field experts, journal articles, press releases, and social media. One of the main hurdles is making sense of complex terms and information when first encountering a potentially newsworthy scientific article or preprint. Even experienced reporters may struggle with scientific jargon that requires significant time and effort to comprehend within a given article's context. While press releases can aid understanding, they may also exaggerate results \cite{sumnerExaggerationsCaveatsPress2016}. Field experts can help simplify and verify information but may be time-constrained \cite{besleyUnderstandingScientistsWillingness2018}, making it difficult for reporters to involve them early in the story identification process. In fact, recent work has distinctly surfaced the need for tools to support sense-making of complex information, especially for science journalists engaged in newsgathering \cite{maidenAutomatingScienceJournalism2023, nishalUnderstandingPracticesComputational2024}.

Generative AI models, such as large language models (LLMs), present a potential solution to this challenge. These models can generate plausible text from user-provided prompts, and techniques like Retrieval-Augmented Generation (RAG) enable generated text to be grounded in the source material \cite{lewisRetrievalAugmentedGenerationKnowledgeIntensive2020}, without the need for domain-specific training data or fine-tuning. Instead, RAG relies on an extra step of knowledge retrieval from the source document during text generation. We hypothesize that a system leveraging these capabilities could help reporters understand scientific jargon more efficiently and reliably.

In this pilot study, we conduct an initial evaluation of such a system that leverages LLMs with RAG to assist readers in understanding jargon within the context of a potentially newsworthy scientific article. All abstracts and jargon terms are in the English language. Specifically, we prototype this as a human-in-the-loop system that aims to:

\begin{itemize}

	\item  Identify complex jargon from scientific abstracts based on the reader's self-reported knowledge, personalizing the results.
	\item Generate a short, accessible definition for each jargon term.
 
\end{itemize}

OpenAI's GPT-4 is used to identify jargon terms and generate definitions, and we specifically use RAG to ground the generated definitions in the context of the abstract or fulltext of the scientific article. The prompts are also designed to personalize the identification jargon terms, based on the subject-matter expertise of a reader. Our approach is part of a larger trend in identifying and defining jargon, often by means of personalization or via novel language models, but most prior work has been directed toward scientists or lay-readers \cite{guoPersonalizedJargonIdentification2023, kimSimpleScienceLexicalSimplification2016, headAugmentingScientificPapers2021, rakedzonAutomaticJargonIdentifier2017}. The purpose of our prototype is to reflect on the potential of using generative AI for science journalists in particular, and to inform the development of a full-fledged system for these users in the future. 

In our evaluation study, we sample arXiv preprints (n=64) to assess the performance of our LLM-based system. We measure the count, precision, and recall of the personalized jargon terms identified by the LLM, comparing them to jargon terms manually annotated by the first two authors based on their experience reading and understanding the preprints. The annotators also evaluate the accuracy and quality of the LLM-generated jargon definitions by conducting pairwise comparisons between two methods of providing background context to GPT-4 -- using RAG to retrieve context from fulltext, versus simply relying on the abstract to provide context. To illustrate the envisioned use case for this technology, we develop a basic user interface that links scientific abstracts to their respective jargon definitions.

The findings offer preliminary insights into how GPT-4 identifies, personalizes, and defines jargon terms. The model tends to overpredict jargon terms regardless of the reader's expertise (i.e., of the annotators in the study), but offers fairly high recall of the actual jargon terms the readers identified (median recall=0.68). It also preserves the relative differences in how the readers identify jargon, suggesting that personalization is a viable pursuit for LLM tools to support sense-making. For jargon definitions themselves, GPT-4 using only the abstract performs slightly better than GPT-4 with RAG-based context from the fulltext in both accuracy and overall quality. The varying patterns in which approach tends to appeal to which reader indicates that readers' subject-matter expertise may influence how they perceive and evaluate definitions. This suggests a need to explore methods to determine when a jargon term might require lookup within the fulltext or not, and based on the level of detail a given reader is looking for (e.g., this may vary depending on the nature of the story being written by a reporter). The choice of relying on RAG vs. abstracts also has repercussions for the computational costs of such tools in newsrooms. 

Based on this initial pilot, we identify questions and directions for future work in the discussion. An external evaluation with professional reporters is necessary to assess the efficacy of this system in practice, but these preliminary insights can inform the design future prototypes for human-centered evaluation, and serve as a resource for other practitioners and researchers working on similar systems in their own newsrooms. By sharing our findings, we aim to contribute to the collective effort of experimenting with generative AI tools that support journalism. Our code and the relevant data can be found online\footnote{\url{https://github.com/ericlee878/ScienceDeJargonizer}}. 

\section{Methods}

This section describes our approach to collecting data from arXiv, personalizing jargon identification, generating jargon definitions, and evaluating the results. 

\subsection{Data Collection}

We limited our focus to de-jargonizing scientific abstracts in arXiv Computer Science\footnote{\url{https://arxiv.org/archive/cs}} published or updated in March 2024, and from three sub-categories: Artificial Intelligence (cs.AI), Human-Computer Interaction (cs.HC), and Computers and Society (cs.CY). Given the current popularity of these sub-categories in both public discourse and scientific research, we believed they could provide illustrative examples to evaluate the proposed RAG system. To this we added another constraint: we collected all peer-reviewed preprints\footnote{This is self-reported by authors in the \mintinline{latex}{comments} field in arXiv metadata.}, which were likely to be fairly credible and well-written. This surfaced a set of 254 articles, mostly from cs.HC (116) or cs.AI (102), with a handful from cs.CY (36). To maintain this sub-category distribution while accommodating resource constraints in terms of availability of annotators, we sampled 25\% of this dataset, resulting in a final test sample of 64 articles (cs.HC: 29, cs.AI: 26, cs.CY: 9). Preprints not included in our final sample were maintained separately as a development corpus for refining LLM prompts and parameters.

\subsection{Personalizing Jargon Identification}

For the test sample, the two annotators independently annotated 5 abstracts for jargon terms that were difficult to understand or infer from the context and generally impeded their comprehension of the work. Subsequently, we engaged in a discussion to establish a shared understanding of what constituted jargon for each annotator. The annotators agreed to rely solely on the abstracts for identifying jargon, and allowed for n-grams to identify complex concepts that were hard to understand (e.g., annotating "Lagrangian-guided Monte Carlo tree search" instead of just "Lagrangian" or "Monte Carlo"). The annotators will be referred to as \mintinline{latex}{rid0} and \mintinline{latex}{rid1} for the rest of this paper, based on the IDs allocated to them during data analysis.

We used OpenAI's GPT-4 to identify jargon terms tailored to the annotators' self-reported scientific expertise in AI, HCI, and Computer Science. Personalizing jargon identification based on readers' varying levels of scientific knowledge can enhance the reading experience and help readers manage their attention more effectively. Prior research on using LLMs and in-context learning to personalize jargon identification for scientists showed that providing metadata about readers' self-identified areas of expertise, citation counts, year of first published paper, and other details could successfully aid in identifying jargon terms \cite{guoPersonalizedJargonIdentification2023}.

We adapted these findings into prompt templates suitable for non-scientists by describing annotators' experiences (e.g., "regularly reads and writes about …", "has taken university classes in …") and self-reported ratings of expertise (e.g., "3/5, where 5 implies …") before prompting the model to identify jargon in a given abstract. Informal pilots for refining prompts and parameters showed that describing experiences worked more consistently than using ratings or a combination of the two. For future iterations with reporters, we will also explore providing more granular metadata to the model, such as random samples of their written stories, bookmarked and/or read scientific papers, or systematically solicited descriptions of topical knowledge, to further improve personalization. The annotator descriptions and model parameters for this task can be found in Appendix \ref{appendex:jargon_id}.

GPT-4's suggestions were validated against human annotations using precision, recall, and F2 scores. F2 scores emphasize recall over precision, and were chosen because we believed that higher recall was preferable for an enhanced reading experience, i.e., identifying as many jargon terms as possible from the ground-truth and reducing false negatives.

\subsection{Generating Jargon Definitions}

We used OpenAI's GPT-4, reinforced with Retrieval-Augmented Generation (RAG)\footnote{Setup using the \mintinline{latex}{llama-index} framework: \url{https://docs.llamaindex.ai/en/stable/}} to generate definitions of jargon terms identified by the human annotators. RAG improves language models by integrating a knowledge retrieval step during text generation. It finds contextually relevant snippets from an external corpus (e.g., the fulltext of an arXiv CS preprint) using cosine similarity, augments the input prompt (e.g., the jargon term to be defined) with the retrieved information, and generates responses grounded in that context with the goal of producing more accurate and informed responses.

We tested different system prompts for the RAG-based approach; query prompts for definitions; and cosine similarity thresholds for retrieving jargon terms from the fulltext in RAG. 
We chose a cosine similarity threshold of 0.3 for the retrieval step in RAG to surface a wider range of relevant chunks, while still excluding completely irrelevant text. For a jargon term like "Lagrangian-guided Monte Carlo tree search", this threshold allowed for the retrieval of chunks containing "Lagrangian", "Monte Carlo", or "tree search", as well as slightly modified versions such as "Monte Carlo search" or "Lagrangian-guided search". We also computed baseline jargon definitions using only the abstract as context, in contrast to the RAG approach that retrieves context from the fulltext. This allows us to compare the effectiveness and computational efficiency of using limited context versus the entire preprint for generating jargon definitions\footnote{We also piloted GPT-4 with no context and found its performance to be significantly worse, making a full evaluation unnecessary.}. Appendix \ref{appendex:jargon_defs} lists the relevant model parameters.

Drawing from prior work that highlights the higher consistency of pairwise comparisons (relative to absolute ratings) for evaluating the outputs of text generation systems \cite{novikovaRankMEReliableHuman2018}, and popular leaderboards like Chatbot Arena \cite{chiang2024chatbot} that implement these recommendations, we evaluated the two approaches using a combination of absolute and relative metrics.

Annotators first assessed each definition's \textbf{individual accuracy} based on the claims made in the context of a given paper. They then evaluated the \textbf{relative quality} of each definition through pairwise comparisons, considering its understandability and usefulness in interpreting the jargon term within the abstract. For each jargon term, annotators expressed their preference between the RAG-generated and abstract-only definitions, with the option to indicate a tie. Sources of the definitions were anonymized and the order was randomized to prevent biases.

We calculated the percentage of correct and incorrect definitions for each approach to gauge their overall accuracy. We also determined the percentage of "wins" for each method in the pairwise quality comparisons, similar to the Elo rating system used in the aforementioned Chatbot Arena. This win percentage provides insight into the relative performance of the two approaches, indicating which one tends to generate more preferred definitions in terms of quality.

\section{Findings}

\subsection{Personalizing Jargon Identification}

\begin{figure}[t]
  \centering
  \includegraphics[width={1.0\linewidth}]{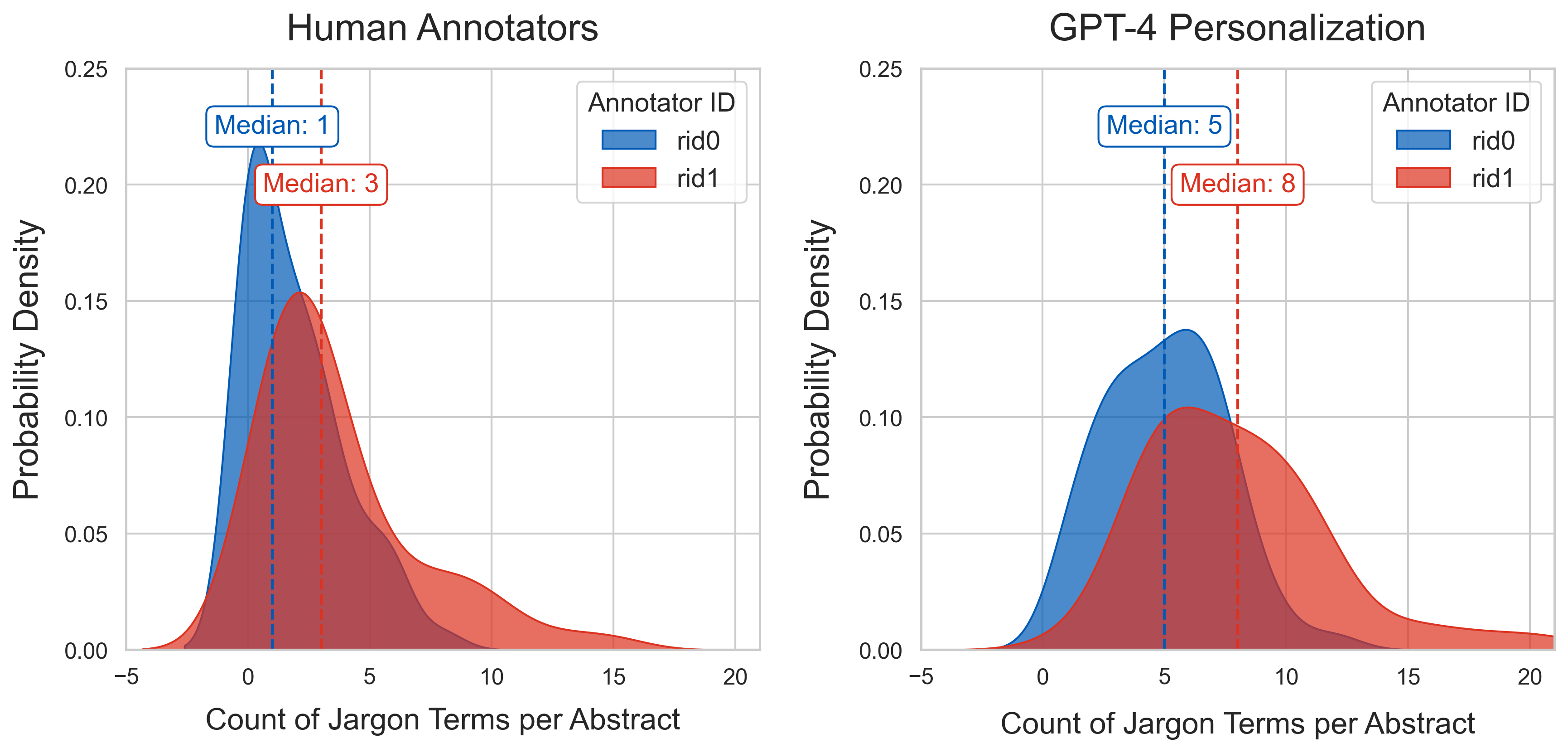}
  \caption{Distribution of jargon terms identified per abstract by human annotators (left) and GPT-4 (right). The probability density plots illustrate how frequently each count of jargon terms is observed. While GPT-4 tends to overestimate the number of jargon terms, it still captures the relative differences between annotators.}
  \Description[Jargon term identification by human annotators vs. GPT-4.]{The left panel shows the distribution of jargon terms identified per abstract by two human annotators. The right panel displays the same for GPT-4. GPT-4 overestimates the number of jargon terms but maintains the relative trend between annotators in our pilot study.}
  \label{fig:humanvsgpt_kdeplot}
\end{figure}

We find that \textbf{GPT-4 consistently identifies a higher number of jargon terms compared to human annotators}, with a median of 4 extra terms for \mintinline{latex}{rid0} and 5 extra terms for \mintinline{latex}{rid1}. Wilcoxon signed-rank tests reveal that these differences are significant (p < 0.01) across the entire dataset and for the individual annotators. 

One might question if this difference stems from lexical variation in how jargon is identified, i.e., if each individual jargon term identified by a human annotator encompasses more words than those identified by GPT-4. However, \textbf{the actual number of words within the jargon terms identified by human annotators and GPT-4 are not largely different}. In fact, GPT-4 identifies slightly longer jargon terms, with a mean difference of 0.3 more words per jargon term compared to human annotators (Mann-Whitney U test; p < 0.01), suggesting that the difference in jargon term count is not due to a systematic difference in jargon term length. Instead, GPT-4 simply tends to suggest a high number of jargon terms, further evidenced by the fact that \textbf{it never predicts an empty set of jargon terms for any abstract}, whereas the human annotators occasionally do so. While this overprediction may create a more noisy reading experience for advanced readers, it may potentially benefit less experienced reporters who are still learning about the technical specifics of the domains they cover.

Figure \ref{fig:humanvsgpt_kdeplot} shows the probability density of jargon terms identified per abstract by human annotators (left panel) and GPT-4 (right panel). The probability density represents the likelihood of observing a certain number of jargon terms across different abstracts, providing insight into the distribution rather than just raw counts. For example, the peaks around 1 (blue, \mintinline{latex}{rid0}) and 3 (red, \mintinline{latex}{rid1}) in the left panel indicate that most human annotations cluster around those values for each annotator. Despite GPT-4 tending to overestimate the count for both annotators, as seen in the right panel, it still maintains the relative difference between annotators. Specifically, \mintinline{latex}{rid1} (red) consistently identifies more jargon terms than \mintinline{latex}{rid0} (blue), which is reflected in both human and GPT-4 predictions. This supports the idea that GPT-4 could capture individual patterns of term identification based on personalized prompting.

Moreover, when human annotators do identify jargon terms, GPT-4 achieves a high median recall of 0.68 (0.83 for \mintinline{latex}{rid0} and 0.66 for \mintinline{latex}{rid1}), and thus \textbf{successfully captures a substantial proportion of the human-identified jargon terms}. However, as expected based on the patterns of over-prediction, the precision is substantially lower for both annotators at a median of 0.33 (0.33 for \mintinline{latex}{rid0} and 0.27 for \mintinline{latex}{rid1}). This results in a median F2 score of 0.55 (0.57 for \mintinline{latex}{rid0} and 0.53 for \mintinline{latex}{rid1}). 
In cases of very low precision (<=0.15, n=20), we find that GPT-4 identifies more jargon than its median, while human annotators identify less, but no clear patterns emerge from qualitative examination of the abstract and jargon terms themselves. A larger sample may be necessary to discern the causes of this mismatch.


\subsection{Generating Jargon Definitions}

We evaluated the accuracy of RAG-based and abstract-based definitions for jargon terms identified by two annotators (n=121 for \mintinline{latex}{rid0}, n=236 for \mintinline{latex}{rid1}). Since accuracy is an objective metric and not dependent on individual annotators, we examined it at the aggregate-level across both annotators. 

Surprisingly, \textbf{the abstract-based approach achieved a slightly higher percentage of accurate responses} at 96.6\%, compared to the RAG-based approach, which had an accuracy of 93.5\%. This is counter-intuitive, as one might expect the RAG-based approach, which draws and contextualizes from the fulltext, to provide more accurate definitions. However, it seems that abstracts contain sufficient context to accurately define jargon terms, while the additional information from the fulltext may introduce noise or irrelevant details, slightly reducing the RAG-based approach's accuracy. 

The RAG-based approach also occasionally failed to provide definitions for some terms (2.5\%), when no snippets of the fulltext met the cosine similarity threshold (0.2) in the retrieval step. This \textbf{thresholding strategy, intended to reduce inaccurate responses, did not seem to prove as effective}.

The 3-4\% inaccuracy in both approaches (n=12 for abstract-based; n=14 for RAG-based) is concerning. For n=7 terms, both approaches return incorrect responses, misinterpreting specific terms as general concepts (e.g., "CodeContests", a benchmark dataset, incorrectly explained to refer to "code contests" as an event). This suggests a need for mechanisms to override GPT-4's tendency to rely on latent word associations and focus on the context provided by the paper, especially when RAG actually surfaces pertinent snippets.

\begin{table}
    \caption{Win, loss, and tie percentages in quality of the RAG-based definitions vs. the abstract-based definitions. Higher win \%-ages are bolded.}
    \Description[Win, loss, tie percentages]{A table with a list of win, loss, and tie percentages of RAG-based approach vs. the abstract-based approach. Higher win \%-ages in bold.}
    \label{table-wins-losses}
    \begin{tabulary}{\linewidth}{llrrr}
        \toprule
        \textbf{Annotator ID}  & Model & \textbf{Win \%}   & \textbf{Loss \%}  & \textbf{Tie \%}  \\
        \midrule
        rid0   & RAG-based & 38.6 \%   & 37.7 \%    & 23.7 \% \\
               & Abstract-based & \textbf{39.4 \%}   & 35.8 \%   & 24.8 \%\\
        \hline
        rid1   & RAG-based & 22.0 \%   & 22.9 \%   & 55.0  \% \\
               & Abstract-based & \textbf{24.1 \%}   & 21.4 \%   & 54.5 \%\\
        \hline
        Overall  & RAG-based & 27.8 \%   & 28.0 \%  & 44.3 \% \\
                 & Abstract-based & \textbf{29.2\%} & 21.4 \% & 54.5 \% \\
        \hline
    \end{tabulary} 
\end{table}

The quality evaluation also reveals a more nuanced picture than expected. Table \ref{table-wins-losses} shows that the abstract-based model slightly outperforms the RAG-based model for both annotators (by 1-2 percentage points in win percentage), suggesting that \textbf{using only the abstract might be as effective as using the fulltext with a RAG approach for generating informative definitions}. This is a little counter-intuitive, as one would expect the extra context around a jargon term to contribute to its informativeness. Again, the potential noise introduced through RAG might be reducing quality, but a fuller, qualitative evaluation is needed to confirm this. 

Interestingly, the high percentage of ties for rid1 (54-55\%) compared to rid0 (24-25\%) also suggests that \textbf{individual differences in subject matter familiarity may influence evaluation criteria}. This may also apply to science reporters with varying levels of expertise, who might find different value in abstract-based or RAG-based (and potentially more dense) definitions. Further qualitative evaluation is necessary to understand these differences tangibly.

\begin{figure}[t]
  \centering
  \includegraphics[width={1.0\linewidth}]{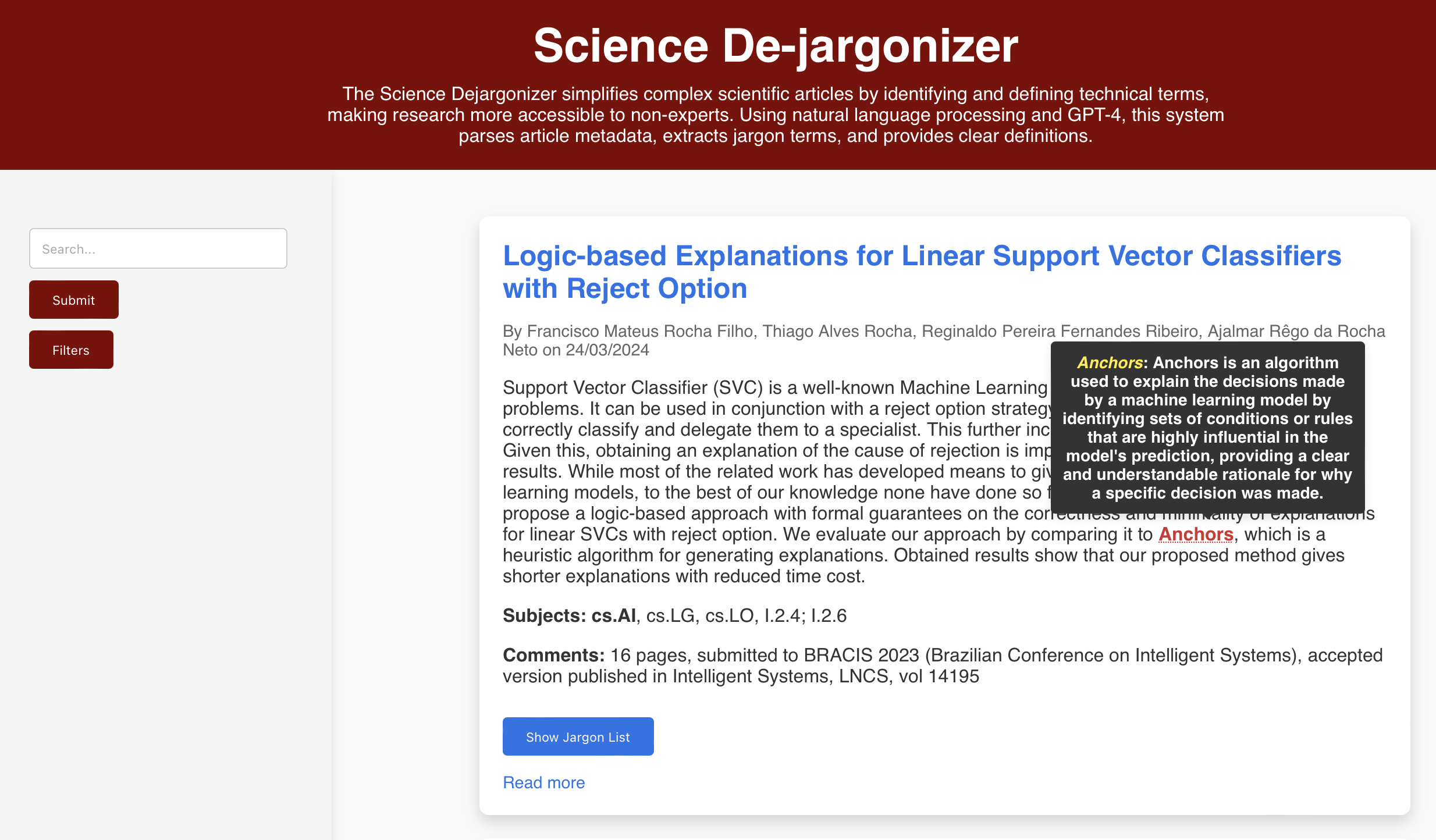}
  \caption{Prototype UI displaying a search bar, filter options, and preprint abstract metadata. Users can hover over specific jargon terms, or scroll through a clickable list to see definitions. Readers can explore the interactive version of the prototype via the linked GitHub repository.}
  \Description[Screenshot of prototype UI.]{Prototype UI displaying a search bar, filter options, and preprint abstract metadata. Jargon terms are specified at the bottom.}
  \label{fig:interface}
\end{figure}

Figure \ref{fig:interface} also shows a prototype user interface (UI) for this system, wherein a user selects arXiv CS sub-categories of interest, and optionally enters some search terms, to then browse through the relevant set of abstracts, with jargon terms personalized to the user and definitions offered below the abstract for easy reference.

\section{Discussion and Conclusion}

In summary, this analysis offers preliminary insight into how GPT-4 identifies, personalizes, and defines jargon terms. While this analysis was conducted within a lab setting with a small number of samples and annotators, we believe some of our findings may be useful for future iterations of this work, as well as for others seeking to build similar tools to simplify information-dense documents for the benefit of reporters.

For one, the model tends to overpredict jargon terms, no matter the reader's expertise, and in fact never returns an empty response. This may not be as much of an issue for reporters as it may be for scientists and researchers (as \mintinline{latex}{rid0} was in this study), but mechanisms to build a more precise personalization system, such as by offering more metadata about the reader or with prompt engineering to encourage empty responses where necessary, can still create a smoother, less noisy reading experience for everyone. The fact that the model still seems to preserve the relative differences in how the two annotators identify jargon, combined with the moderate to high recall for each annotator, signals that personalization is a viable pursuit. A more extensive evaluation with practicing reporters as well as functionality to adapt the LLM's internal user model based on feedback can lead to better, more tailored systems in the future.

Inaccuracies remain an ongoing concern, at a 3-4\% rate across both tested approaches. On the one-hand, we can argue that verifying information past the point of news discovery is part of the process of science reporting, and that this may be analogous to other claims that reporters need to fact-check (e.g., claims from press releases and company blogposts). On the other hand however, we do want these tools to actually streamline the reading and verification experience during news discovery, and counter the challenges of existing approaches. Potential solutions here range from prompt and parameter engineering (e.g., reducing \mintinline{latex}{temperature} parameter to 0), to calibrating user expectations around the performance of LLMs via specific disclaimers, to actually implementing interface-based interventions (e.g., highlighting the uncertainty of words or tokens in LLM-generated text) as heuristics for reporters to decide what they need to focus their attention on verifying. Empirical studies are needed to test the effect of such approaches on the efficiency of the reading and sense-making experience.

The quality of the generated definitions also paints a complex picture: GPT-4 using only the abstract performs slightly better than GPT-4 with RAG-based context from the fulltext, both in accuracy and overall quality. One possible explanation is that the RAG-based approach may introduce noise by surfacing less relevant information from the fulltext. However, it is still unexpected that the abstract-based approach can generate comparable responses for niche jargon terms, which are not always clearly defined in the abstract alone. This raises questions about the source of information used by the abstract-based model, as the human annotators often needed to search the fulltext to verify the accuracy of the generated definitions. We contend it could be an effect of GPT-4's extensive size and training data -- a comparison of approaches with smaller (and also more cost-effective) models may help clarify this. If the model indeed relies more on prior knowledge from training data rather than provided context, it may be unable to accurately define jargon terms that emerged after the cutoff date for LLM model pretraining, without fulltext context like from RAG.

The varying tie percentages for both annotators in Table \ref{table-wins-losses} also suggest that different aspects of the RAG or abstract-based definitions may appeal to readers differently, depending on their subject-matter expertise. Another hypothesis, albeit without statistical backing, is that some terms may benefit from a definition strung together from different parts of the abstract, or from latent associations encoded in LLM weights during training, while others may require more detailed or specific context from the fulltext. This may depend on a reader's expertise, but also on a jargon term's importance to the paper's contribution and findings. For example, in a paper about reducing the security vulnerabilities of some hardware design plans, jargon referring to a programming language in reference to research methods (e.g., "SystemVerilog") might require less explanation than a conceptual jargon term about the research question itself (e.g.,"Common Weakness Enumerations")\footnote{Both are real terms in our dataset, sourced from \url{https://arxiv.org/abs/2403.16750}.}. Future work could involve developing heuristics to determine when a jargon term might require lookup within the fulltext versus not, based on the reader, on how important the term is to an article's contribution, and how much detail the reader is interested in exploring (e.g., this may also vary depending on feature-length vs. study stories).

\bibliographystyle{ACM-Reference-Format}
\bibliography{sample-base}

\appendix

\section{Model Parameters}

\subsection{Personalizing Jargon Identification}
\label{appendex:jargon_id}

The following model parameters were used: \mintinline{latex}{gpt-4-turbo} model (2024-05-25), \mintinline{latex}{max_tokens} = 512, \mintinline{latex}{temperature} = 1. Table \ref{table-annotators} shows annotator descriptions used to personalize jargon identification:

\begin{table}[H]
    \scriptsize
    \caption{Annotator Descriptions}
    \Description[Annotator Descriptions]{A table with a list of annotators and their level of expertise, self-described}
    \label{table-annotators}
    \begin{tabulary}{\linewidth}{LL}
        \toprule
        \textbf{ID} & \textbf{Annotator Description}  \\ 
        \midrule
        rid0 & PhD student and researcher in Human-Computer Interaction and Artificial Intelligence. The reader has a few papers in top-tier conferences where they used their knowledge of Natural Language Processing, Machine Learning, HCI and software engineering to build and evaluate interactive systems. \\
        \midrule
        rid1 & Sophomore in Computer Science, with experience in Software Engineering. The reader has taken a few courses and worked on some class projects in Machine Learning and Software Engineering, and is interested in learning more about the fields. \\
        \bottomrule
    \end{tabulary} 
\end{table}

System and query prompts can be found in our GitHub repository.





\subsection{Generating Jargon Definitions}
\label{appendex:jargon_defs}

OpenAI's \mintinline{latex}{text-embedding-3-small} model was used to retrieve relevant snippets for RAG. The following parameters were used to generate definitions: \mintinline{latex}{gpt-4-turbo} model (2024-05-25), \mintinline{latex}{max_tokens} = 512, \mintinline{latex}{temperature} = 1. 

System and query prompts can be found in our GitHub repository.






\end{document}